\documentclass[12pt, a4paper]{article}
\usepackage[utf8]{inputenc}
\usepackage[pdftex]{graphicx}
\usepackage[lofdepth,lotdepth,labelformat=empty]{subfig}
\usepackage{endnotes}
\let\footnote=\endnote
\usepackage{rotating}
\usepackage{pslatex}
\usepackage{colortbl}
\usepackage[margin=1in]{geometry}
\usepackage{setspace}
\usepackage{lscape}
\usepackage{multirow}
\usepackage{ctable}
\usepackage{amsmath}
\usepackage{natbib}
\usepackage{color}
\usepackage{eurosym}
\usepackage{booktabs}
\usepackage{dcolumn}
\newcolumntype{.}{D{.}{.}{-1}}
\usepackage[toc,page]{appendix}

\newcolumntype{C}[1]{>{\centering\let\\\tabularnewline}p{#1}}
\newcolumntype{R}[1]{>{\raggedleft\let\\\tabularnewline}p{#1}}
\newcolumntype{L}[1]{>{\raggedright\let\\\tabularnewline}p{#1}}

\usepackage{color}
\definecolor{darkred}{rgb}{0.5,0,0}
\definecolor{darkgreen}{rgb}{0,0.5,0}
\definecolor{darkblue}{rgb}{0,0,0.5}

\usepackage{hyperref}
\hypersetup{colorlinks,
 linkcolor=black,
 filecolor=darkgreen,
 urlcolor=darkblue,
 citecolor=darkblue,
 pdftitle={UNGDC},
 pdfauthor={Alexander Baturo, Niheer Dasandi, and Slava Mikhaylov},
 pdfsubject={UN General Debates Corpus}
}

\begin{document}

\title{Understanding State Preferences With Text As Data: Introducing the UN General Debate Corpus
\thanks{Authors' names are listed in alphabetical order. Authors have contributed equally to all work. We acknowledge the receipt of Dublin City University Enhancing Performance Award. We would like to thank Sofia Collignon Delmar, Elvin Gjevori, Karl Murphy, Mohsen Moheimany, and Bethsabee Souris for their excellent research assistance. We are also grateful to Kristin Bakke, Alex Braithwaite, David Hudson, Tim Hicks, Jeff Kucik, Lucas Leemann, Neil Mitchell, and Erik Voeten for their helpful comments and advice. UN General Debate Corpus is made available on the Harvard Dataverse at \url{http://dx.doi.org/10.7910/DVN/0TJX8Y}.}
}

\vspace{-0.5in}
\author{
Alexander Baturo\\
  Dublin City University\\
  \href{mailto:alex.baturo@dcu.ie}{alex.baturo@dcu.ie}
\and
Niheer Dasandi\\
  University of Birmingham\\
  \href{mailto:n.dasandi@bham.ac.uk}{n.dasandi@bham.ac.uk}
\and
Slava J. Mikhaylov \\
 University of Essex\\
  \href{mailto:s.mikhaylov@essex.ac.uk}{s.mikhaylov@essex.ac.uk}\\  
 }

\date{1 July 2017}


\maketitle
\begin{abstract}
\begin{spacing}{1.5}
\noindent  \normalsize Every year at the United Nations, member states deliver statements during the General Debate discussing major issues in world politics. These speeches provide invaluable information on governments' perspectives and preferences on a wide range of issues, but have largely been overlooked in the study of international politics. This paper introduces a new dataset consisting of over 7,300 country statements from 1970--2014. We demonstrate how the UN General Debate Corpus (UNGDC) can be used to derive country positions on different policy dimensions using text analytic methods. The paper provides applications of these estimates, demonstrating the contribution the UNGDC can make to the study of international politics.  
\end{spacing}

\vspace{.5cm} \noindent \textbf{Key Words}: Policy preferences, foreign policy, United Nations, text as data

\end{abstract}

\thispagestyle{empty}

\newpage
\begin{spacing}{2}

\subsection*{Introduction}

Every September, the heads of state and other high-level country representatives gather in New York at the start of a new session of the United Nations General Assembly (UNGA) and address the Assembly in the General Debate. The General Debate (GD) provides the governments of the almost two hundred UN member states with an opportunity to present their views on international conflict and cooperation, terrorism, development, climate change, and other key issues in international politics. As such, the statements made during GD are an invaluable and, largely untapped, source of information on governments' policy preferences across a wide range of issues over time.

Government preferences are central to the study of international relations and comparative politics. As preferences cannot be directly observed, they must be inferred from states' observed behavior. One approach has been to use military alliances as an indicator of preference similarity \citep[e.g.][]{de1983war}. This approach, however, provides little information about preferences when states do not have alliances. Scholars have instead overwhelmingly relied on UNGA voting records to estimate foreign policy preferences \citep[see][]{voeten2013data,bailey2013}. However, UNGA voting-based methods -- like all measures of preference -- rely on certain assumptions, and as such, have both strengths and limitations \citep[see][]{voeten2013data}. For example, one shortcoming is estimates of state preference are derived from the limited number of issues that are voted on in the UNGA in a given year.\footnote{\normalsize As \citet{hage2012consensus} explain, the UNGA has always practiced the adoption of many items on its agenda without a formal vote.} Therefore, it is essential that researchers can draw on additional data and measures to avoid producing findings about government preferences that are based on one type of observed state behavior. 

We argue that the application of text analytic methods to General Debate statements can provide much-needed additional measures and tools that can broaden our understanding of government preferences and their effects. The use of text analytic methods is rapidly gaining ground in comparative politics and legislative studies \citep[see][]{laver:2003,proksch2010position,herzog2015}. To date, however, there has been little effort to use speeches to estimate policy preferences in international relations. The formal and institutionalized setting of the GD; its inclusion of all UN member states -- which are provided with equal opportunity to address the Assembly; and the fact that it takes place every year, makes the GD ideal for using text analysis to derive estimations of state preferences that can be applied to systematic analyses of international politics.

This paper introduces a new dataset, the UN General Debate Corpus (UNGDC), consisting of 7,314 General Debate statements from 1970-2014, that we have pre-processed, categorised and prepared for empirical applications. In the next section, we discuss the characteristics, content, and purpose of the UN General Debate. Secondly, we explain the process of collecting and preprocessing the statements, and provide an overview of the UNGDC. We then use the text as data approach we show how the UNGDC can be used to derive estimates of government preferences, providing applications of these estimates. We conclude by outlining potential uses of the UNGDC in future research.

\subsection*{The UN General Debate and world politics}

The General Debate marks the start of the UNGA regular session each year. By tradition, since 1947, the opening speech is made by Brazil, with the US also scheduled to speak on the first day. Typically, the heads of state and governments speak during the first days of the GD, followed by vice-presidents, deputy prime ministers and foreign ministers, and concluding with the heads of delegation to the UN \citep{nicholas1959,bailey1960,luard1994,smith2006}. While numbers vary across sessions, on average heads of state or government comprise 39\% of speakers; vice-presidents, deputy prime ministers and foreign ministers make up around 54\% of speakers; with country representatives to the UN constituting 7\% of all speakers. 

The GD provides governments with an opportunity to declare, and to have on public record, their official position on various major international events of the past year \citep{smith2006}. In addition, country representatives use the GD venue to present their governments' perspectives on broader underlying issues in international politics. Their speeches frequently deal with issues of mutual concern such as terrorism, nuclear non-proliferation, development and aid, and climate change -- often appealing to the international community to do more to tackle these issues.  For example, in 1995 the US discussed UN reform, non-proliferation, terrorism, money laundering and the narcotics trade in its GD statement. In turn, the UK and France  both drew attention to the challenges of UN peacekeeping, while India discussed terrorism, disarmament, human rights, and concerns about the inability of global institutions, such as the WTO, to address the needs of the Global South. 

There are several important characteristics of GD speeches that have implications for the use of these statements in deriving estimates of state preferences. In contrast to UNGA roll-call votes, which are directly linked to the adoption of UN resolutions, GD speeches are not institutionally connected to decision-making within the UN. As a consequence, states face lower external constraints and pressures when delivering GD statements than when voting in the UNGA. Indeed, studies that use UNGA voting highlight the various constraints countries face when voting as a result of, among other things, aid relationships and strategic voting blocs \citep[see][]{kim:1996,voeten:2000,alesina2000gives}. The lack of external constraints means that when delivering their GD statements, governments have more leverage with the positions they take and the issues they emphasize. Hence, GD statements provide more information on key national priorities than the limited number of votes in the UNGA. 

This view is supported by interviews conducted by the authors with members of the diplomatic community. The Deputy Representative of the Finnish Mission to the UN, for example, explained, ``speeches at the General Debate are interesting because they flesh out national policies -- what states think... the General Debate is the one place where states can speak their mind; it reflects the issues that states consider important." Similarly, a spokesperson for the German Mission to the UN stated that the absence of external pressures when delivering GD statements means ``these speeches are the most sovereign thing that a country does as a member of the UN.''\footnote{\normalsize These quotes are from interviews conducted by Niheer Dasandi and Nicola Chelotti with Jaane Taalas (9 June 2015) and Christian Doktor (16 June 2015).} It is clear that non-democratic regimes also attach great importance to GD statements. For example, members of Russia's inner political circle not only viewed the 2015 GD statement as a key summary of its foreign policy concerns, they were also apparently aware of its content weeks in advance.\footnote{\normalsize Foreign Minister Lavrov revealed the issues Russia planned to discuss during the 2015 General Debate two weeks before the debate. See \url{http://www.interfax.ru/russia/466392}, accessed 27 August 2016.}  

A significant consequence of the relative lack of external constraints in the GD is that member states can more freely express their government's perspectives on issues deemed important -- including more contentious issues. As \citet[155]{smith2006} argues, a key function of the GD is that ``it provides members with the opportunity to blow off steam on contentious issues without causing undue damage.'' This is particularly relevant for smaller nations which can use the GD to raise more disagreeable political issues \citep[see][]{nicholas1959}. For example, in 2014, Antigua and Barbuda's statement emphasized the failure of the US government to adhere to a ruling from the WTO's Dispute Settlement Body that stated that the US should pay compensation to Antigua and Barbuda. In making this complaint, the Antiguan representative highlighted the importance of the GD for smaller nations, stating ``my small nation has no military might, no economic clout. All that we have is membership of the international system as our shield and our voice in this body as our sword.''

The fewer external constraints on representatives when delivering GD statements does not, however, imply that these speeches are not strategic. Scholars have long recognized that ``member states present themselves exclusively in the guise in which they wish to be known'' during these annual debates \citep[p. 98]{nicholas1959}. In fact, a key purpose of the General Debate is that it provides governments with the opportunity to ``influence international perceptions of their state, aiming to position their states favorably, as well as to influence the perception of other states'' \citep[p. 10]{hecht-2016}. Therefore, governments use GD speeches strategically to signal their preferences among the community of states. This use of strategic signaling in the GD can be seen when we compare references to Iran in the US statements in 2012 and 2013. In the 2012 address, President Obama was highly critical of Iran:

\begin{quotation} \noindent In Iran we see where the path of a violent and unaccountable ideology leads [...] Time and again, it has failed to take the opportunity to demonstrate its nuclear program is peaceful [...] Make no mistake: a nuclear-armed Iran is not a challenge that can be contained. It would threaten the elimination of Israel, the security of Gulf nations and the stability of the global economy [...] and that is why the United States will do what we must to prevent Iran from obtaining a nuclear weapon.
\end{quotation}

In contrast, speaking a year later, the US president was more reconciliatory, offering to give diplomacy one last chance in relation to Iran's nuclear program:

\begin{quotation} \noindent ... if we can resolve the issue of Iran's nuclear program, that can serve as a major step down a long road towards a different relationship, one based on mutual interests and mutual respect [...] America prefers to resolve its concerns over Iran's nuclear program peacefully [...] We are not seeking regime change, and we respect the right of the Iranian people to access peaceful nuclear energy [...] 

\end{quotation}

A few hours later during the same session, President Rouhani in his address also emphasized diplomacy and the hope of reaching a compromise. The world has subsequently learned that in the background, the US and Iran held secret talks that eventually led to the breakthrough and signing of the intermediate deal \citep{guardian}. As such, the change in rhetoric between 2012 and 2013 demonstrates the strategic nature of GD speeches.  A further example of both the importance placed by governments on the GD address and its strategic purpose is provided by the Chilcot Inquiry into the UK's role in the Iraq War. The report contains a memo sent by Prime Minister Tony Blair to President George W. Bush, complimenting the US President on the speech delivered in the 2002 General Debate setting out the case for war, ``It was a brilliant speech ... it puts us on exactly the right strategy to get the job done.''\footnote{\normalsize Section 3.4 of \emph{the Iraq Inquiry}, p.187, see \url{http://www.iraqinquiry.org.uk/media/248175/the-report-of-the-iraq-inquiry_section-34.pdf}, accessed 25 January 2017.} Hence, the US speech was seen as part of the US and UK strategy to build support for intervention in Iraq. 

The lack of external constraints on member states in delivering GD statements means that they can use their address to indicate the issues considered most important by devoting more attention to these topics. As governments can choose what issues to discuss or ignore, and how strongly to emphasize certain issues, the GD provides detailed information about a government's \emph{position} on a policy issue, and also the \emph{importance} -- or \emph{salience} -- of an issue for a government. As \citet[155]{smith2006} notes, the General Debate acts ``as a barometer of international opinion on important issues, even those not on the agenda for that particular session.'' The focus on position and salience means that GD speeches can be used to uncover the most important topics that emerge in international politics over time.

\vspace{-0.2in}
\subsection*{UNGDC: The UN General Debate Corpus}

The speeches made in the General Debate are subsequently deposited at the United Nations Dag Hammarskjold Library. However, statements made before 1992 are stored as image copies of typewritten documents. These are of very poor image quality and require additional preprocessing using optical character recognition software. We collected speeches through the dedicated webpages of the individual UNGA General Debates and the UN Bibliographic Information System (UNBIS). 

Speeches are typically delivered in the native language. Based on the rules of the Assembly, all statements are then translated by UN staff into the six official languages of the UN. If a speech was delivered in a language other than English, we use the official English version provided by the UN. Therefore, all of the speeches in the UN General Debate Corpus (UNGDC) are in English. 

The annual sessions are assigned numbers, starting with the 1st session in 1946 up to the most recent 70th Session in 2015. We collected all GD speeches from 1970 (Session 25) to 2014 (Session 69). In total, there are 7,314 country statements delivered from 1970-2014. The number of countries participating in the GD increased from 70 in 1970 to 193 in 2014 in line with the increase in UN membership. Non-member states may also participate in the GD (e.g., the Holy See and Palestine). Several states that previously participated in the GD have ceased to exist. Where possible we linked such states to their legal successor states (e.g., USSR and the Russian Federation). If this was not possible we kept speeches in the data under the country's last known name (e.g. German Democratic Republic). Overall, the corpus contains the GD contributions from 198 countries. On average, speeches contain 123 sentences and 945 unique words.\footnote{\normalsize We make the UNGDC publicly available on the Harvard Dataverse at \url{http://dx.doi.org/10.7910/DVN/0TJX8Y}. We have also developed a browsing tool that allows users to explore individual documents and the topics covered, including the top words that characterize topics, the evolution of topics over time, word distributions across topics, the underlying digitized texts of speeches, and the source PDFs at \url{http://ungd.smikhaylov.net}. The website also enables users to download the UN General Debate Corpus.} 

\begin{table}
\centering
\scriptsize
\begin{center}
\begin{tabular} {@{} l c c c c c @{}}
\toprule  
 Year & UN Membership & GD Statements & Types (mean freq)& Tokens (mean freq)& Sentences (mean freq) \\
         \hline \hline 
1970	&	127	&	70	&	1569	&	8230	&	257	\\
1971	&	132	&	116	&	1336	&	5927	&	230	\\
1972	&	132	&	125	&	1157	&	4895	&	180	\\
1973	&	135	&	120	&	1291	&	5923	&	230	\\
1974	&	138	&	129	&	1093	&	4248	&	191	\\
1975	&	144	&	126	&	1041	&	4280	&	165	\\
1976	&	147	&	134	&	951	&	3720	&	151	\\
1977	&	149	&	140	&	965	&	3452	&	135	\\
1978	&	151	&	141	&	1159	&	4169	&	163	\\
1979	&	152	&	144	&	1220	&	4804	&	200	\\
1980	&	154	&	149	&	1173	&	4663	&	183	\\
1981	&	157	&	145	&	1159	&	4357	&	183	\\
1982	&	157	&	147	&	1134	&	3986	&	151	\\
1983	&	158	&	149	&	1078	&	3669	&	157	\\
1984	&	159	&	150	&	1160	&	3951	&	172	\\
1985	&	159	&	137	&	1142	&	3605	&	113	\\
1986	&	159	&	149	&	895	&	2715	&	85	\\
1987	&	159	&	152	&	922	&	3010	&	102	\\
1988	&	159	&	154	&	985	&	3463	&	124	\\
1989	&	159	&	153	&	1036	&	3365	&	117	\\
1990	&	159	&	156	&	1076	&	3606	&	125	\\
1991	&	166	&	162	&	1086	&	3519	&	127	\\
1992	&	179	&	167	&	932	&	2962	&	103	\\
1993	&	184	&	175	&	1062	&	3433	&	135	\\
1994	&	185	&	178	&	1142	&	4040	&	140	\\
1995	&	185	&	172	&	1255	&	4306	&	168	\\
1996	&	185	&	181	&	1220	&	4149	&	157	\\
1997	&	185	&	176	&	915	&	2659	&	122	\\
1998	&	185	&	181	&	892	&	2749	&	115	\\
1999	&	188	&	181	&	857	&	2567	&	91	\\
2000	&	189	&	178	&	937	&	2677	&	88	\\
2001	&	189	&	189	&	681	&	1925	&	78	\\
2002	&	191	&	188	&	588	&	1465	&	58	\\
2003	&	191	&	189	&	666	&	1761	&	72	\\
2004	&	191	&	192	&	557	&	1400	&	61	\\
2005	&	191	&	185	&	505	&	1311	&	51	\\
2006	&	192	&	193	&	554	&	1393	&	63	\\
2007	&	192	&	191	&	573	&	1392	&	52	\\
2008	&	192	&	192	&	609	&	1498	&	59	\\
2009	&	192	&	193	&	662	&	1754	&	65	\\
2010	&	192	&	189	&	631	&	1668	&	58	\\
2011	&	193	&	193	&	709	&	2097	&	79	\\
2012	&	193	&	194	&	626	&	1671	&	66	\\
2013	&	193	&	192	&	776	&	2306	&	71	\\
2014	&	193	&	193	&	555	&	1451	&	50	\\
\bottomrule
\end{tabular}
\end{center}
\caption{ \emph{UN General Debate Corpus.} \emph{Note:} Descriptive statistics for the UNGDC  containing 7,314 statements delivered by heads of state or their representative from 1970--2014. From 2011, President of the European Commission made a separate statement on behalf of the EU.}
 \label{tab:docs}
\end{table}

Table \ref{tab:docs} provides an overview of the UNGDC. The table shows average frequency of types (unique form of a word), tokens (individual words), and sentences for each individual speech in text corpus. In terms of who delivered the statement, 1,909 (44.3\%) were delivered by heads of state or government (e.g. presidents, prime ministers, kings); 2,126 (49.3\%) by vice-presidents, deputy prime ministers, and foreign ministers; and 276 (6.4\%) by country representative at the UN.\footnote{\normalsize Detailed information is available for sessions 49--69, transcripts from earlier sessions do not provide the same degree of detail regarding the post of the speaker. In the rare cases where the post of the speaker was missing in the transcript for sessions 49--69, we added this information.}

\subsection*{Empirical application: Preferences on single issue dimensions}

The UNGDC can be used by scholars who require easy access to the statements and may want to read a particular text, or compare selected statements. Primarily, however, we envision the UNGDC to be used in quantitative applications looking at the nature, formation, and effects of state preferences in world politics. Treating text as data has a long tradition in political science \citep[for a review, see][]{laver2014measuring}. Since the earlier introduction of text scaling methods to estimate policy positions on dimensions of interest -- such as Wordscore \citep{laver:2003} and Wordfish \citep{slapin:2008} -- the availability and complexity of methods has increased dramatically \citep{grimmer2013,herzog2015}. The majority of such methods are either directly derived from, or can be traced to, the natural language processing literature in computer science and computational linguistics \citep[e.g.][]{lowe2008understanding,Benoit2013}. Wordscore is by far the most popular text scaling method in political science based on a Google Scholar citation count. It is related to the Naive Bayes classifier deployed for text categorization problems \citep{Benoit2013}.

\begin{figure}
\centering

\includegraphics[width=.95\textwidth]{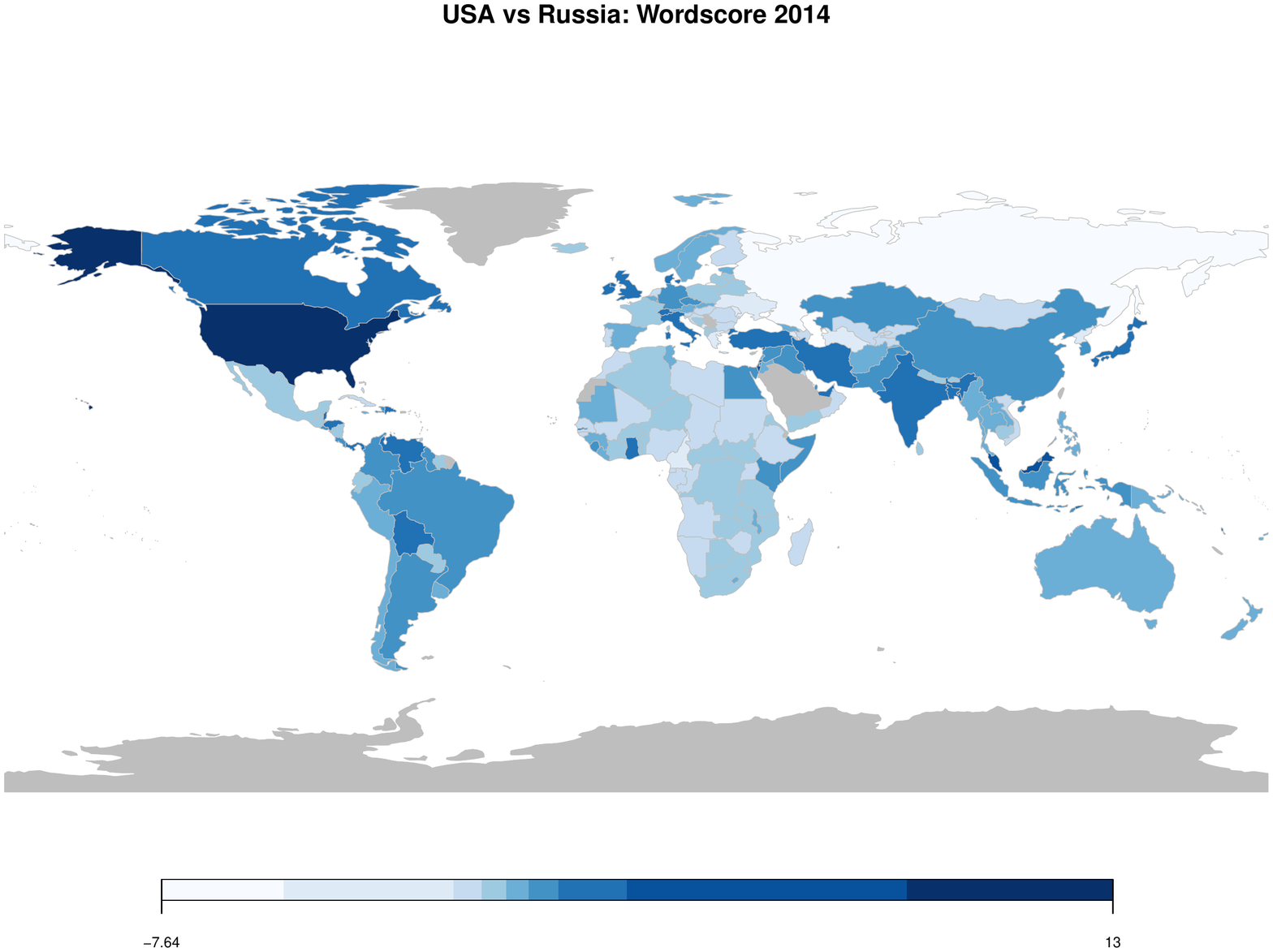} 
\caption{\emph{Wordscore Map 2014}.  \emph{Note:} The scores are estimated in \texttt{quanteda} package (version 0.9.9-3) in R \citep{Benoit2013}. We follow standard preprocessing during the tokenization stage, remove English stopwords, and perform stemming. We also trim the document-feature matrix to have features that appear at least five times in three documents. USA is given reference score ($+$1) and Russia ($-$1). Results are rescaled using classical LBG rescale, hence predicted scores may be beyond the ($-1; +1$) range.
\label{fig:map}}
\end{figure}

Working with text as data generally involves using the bag-of-words approach, whereby each document can be represented by a multiset (bag) of its words that disregards grammar and word order. Word frequencies in the document are then used to classify the document into one of two categories. In Wordscore the learning is supervised by providing training documents that are \emph{a priori} known to belong to either category, so that the chosen dimension is substantively defined by the choice of training documents. 

As an illustration of this approach, we derive estimates of preferences on the very specific issue of USA-Russia rivalry in world politics. Figure \ref{fig:map} maps wordscore estimates for the 2014 UN General Debate. We use statements by the US and Russia as reference texts. We therefore \emph{a priori} define the policy dimension as Russia vs USA. We do not use the resulting scores as an explanatory variable in an empirical application here due to limited space. However, such an application would clearly be of value for IR research. Here, we simply demonstrate how it is possible to derive estimates of differences between UN member states, using the text as data approach.

\subsection*{Empirical application: Preferences on multiple dimensions}

While estimating state preferences on single issue dimensions has many benefits, countries routinely express preferences on multiple dimensions of foreign policy. We therefore turn to correspondence analysis (CA) --- a dimensionality reduction technique \citep[e.g.][]{bonica2013ideology}. In CA, the first dimension is fitted to explain maximal variation in the data, while subsequent dimensions explain maximal residual variation (which means dimensions are orthogonal to each other). Unlike Wordscore, the definition of the dimensions produced by CA must be discerned inductively,  \emph{a posteriori} \citep{laver2014measuring}. This also implies that the dimensions produced by CA may correspond to single, multiple, or meta issues. 

\cite{lowe2016} suggests that position estimated by such models is a low dimensional summary of the relative emphasis of one topic over another, compared to what would be expected by chance. This is consistent with a key assumption of the saliency theory of party competition \citep{budge2001mapping}, which posits that the policy differences between parties are determined by their contrasting emphases on different issues. In the context of GD statements, the CA model fitted to the count data of unique words captures countries' relative emphasis of different issues --- and therefore the differences in their policy preferences. 

A benefit of using CA is that it allows us to easily estimate positions on multiple dimensions. We demonstrate the ease of using multidimensional text scaling by including the new CA measures in an existing analysis of the International Criminal Court (ICC) and US nonsurrender agreements \citep{kelley2007keeps}. The format of this article prevents us from covering issues in detail; therefore the following is intended merely as an illustrative example. In brief, the US sought to pressure other states to sign bilateral agreements not to surrender US citizens to the ICC. This attempt to seek exceptional treatment was widely criticized for inconsistency with international norms, and many countries (but not all) turned it down. \citet[p. 573]{kelley2007keeps} argues that for these states normative preferences trumped strategic concerns. Overall, the views on the nonsurrender agreements were complex and unlikely to be reduced to an easily identifiable single-issue dimension.


\begin{landscape}
\begin{figure}
\centering
\includegraphics[width=.7\textwidth]{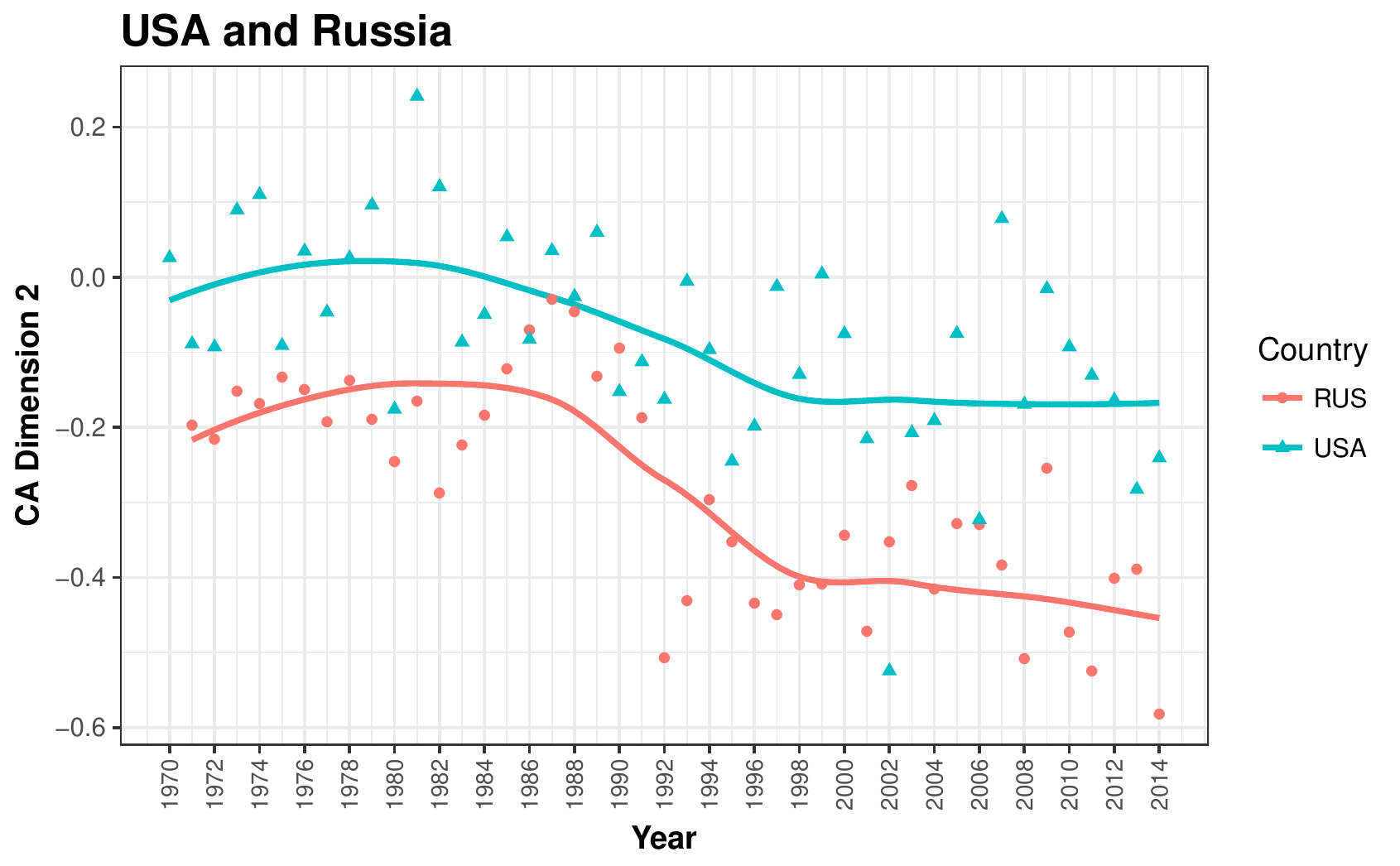} 
\includegraphics[width=.7\textwidth]{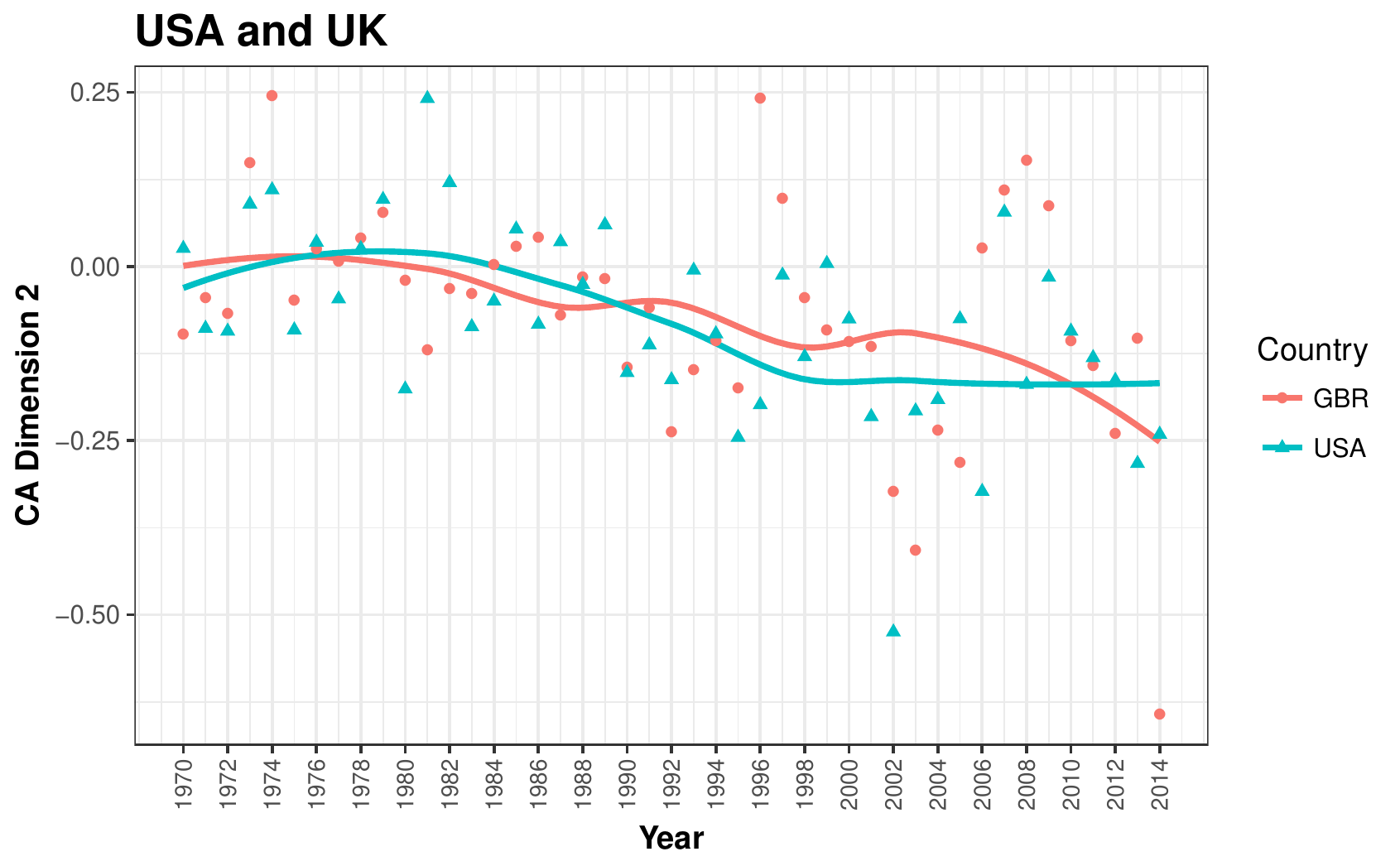} 
\includegraphics[width=.7\textwidth]{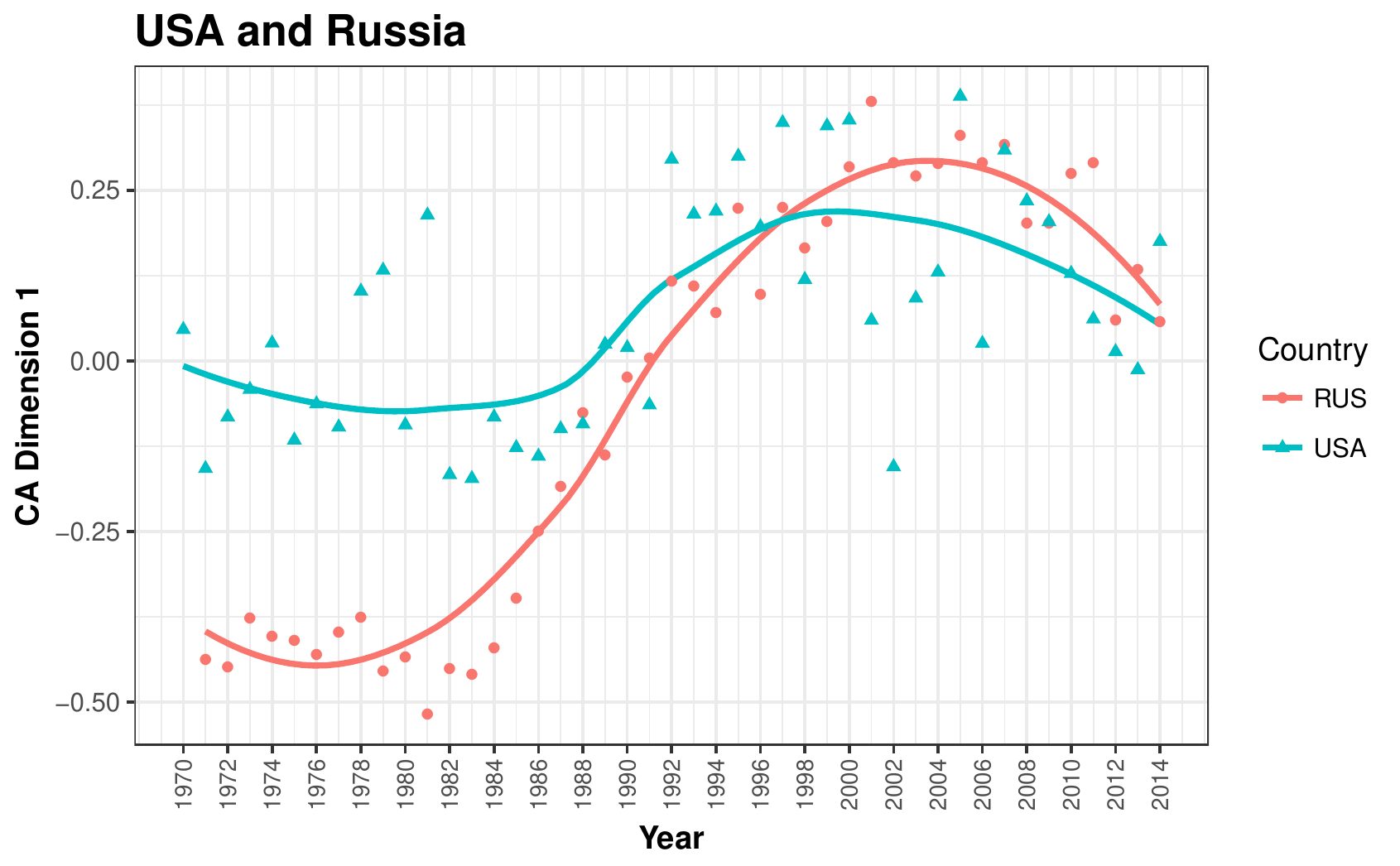} 
\includegraphics[width=.7\textwidth]{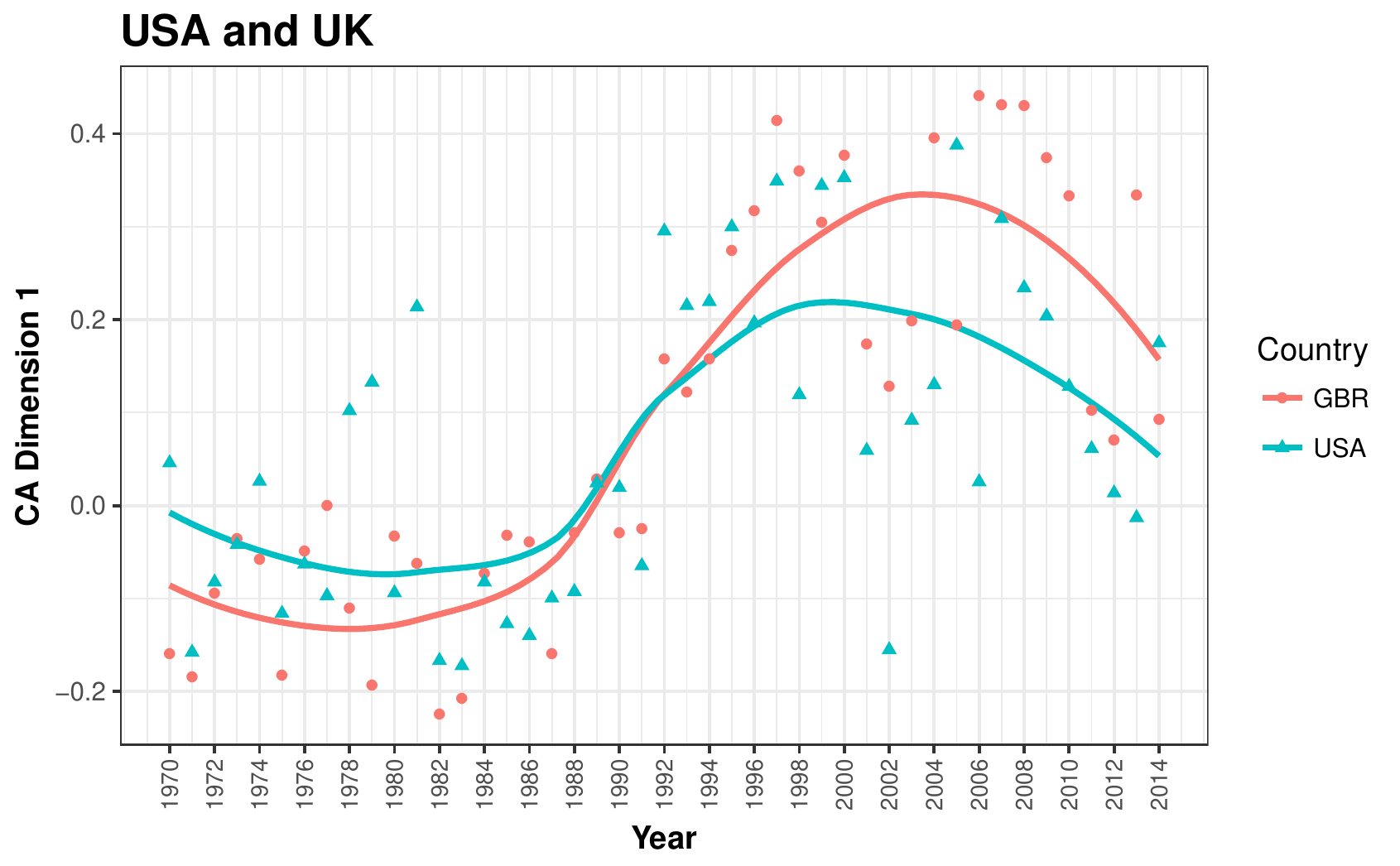}
\caption{\emph{\emph{CA1} and \emph{CA2} of Allies and Opponents}. \emph{Note:} The four subplots show correspondence analysis estimates for USA, Russia and the UK on the first and second dimensions, as discussed in text. Overlaid lines are loess smoothers.
\label{fig:ca1ca2}}
\end{figure}
\end{landscape}

\begin{figure}
\centering

\includegraphics[width=.95\textwidth]{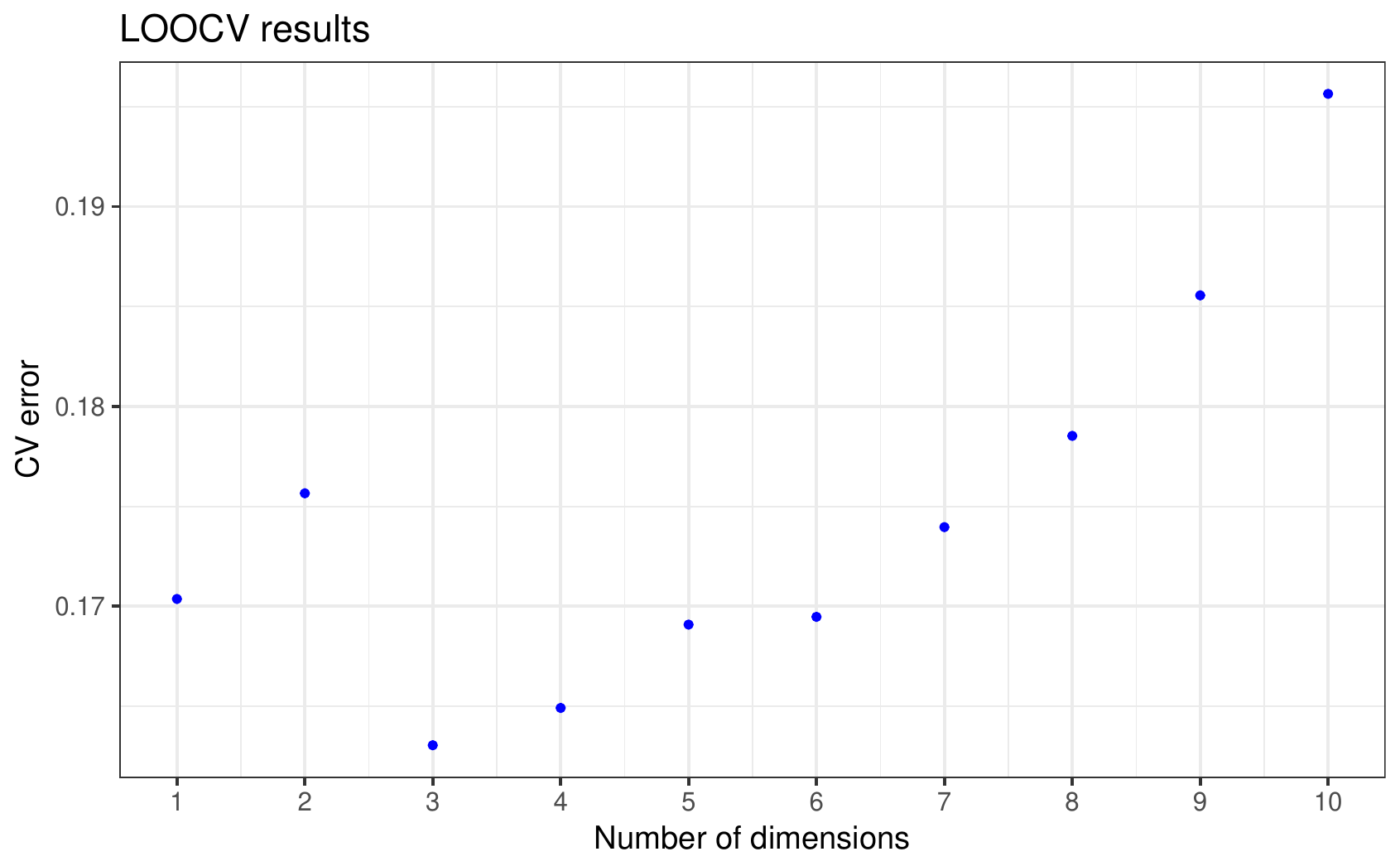}
\includegraphics[width=.95\textwidth]{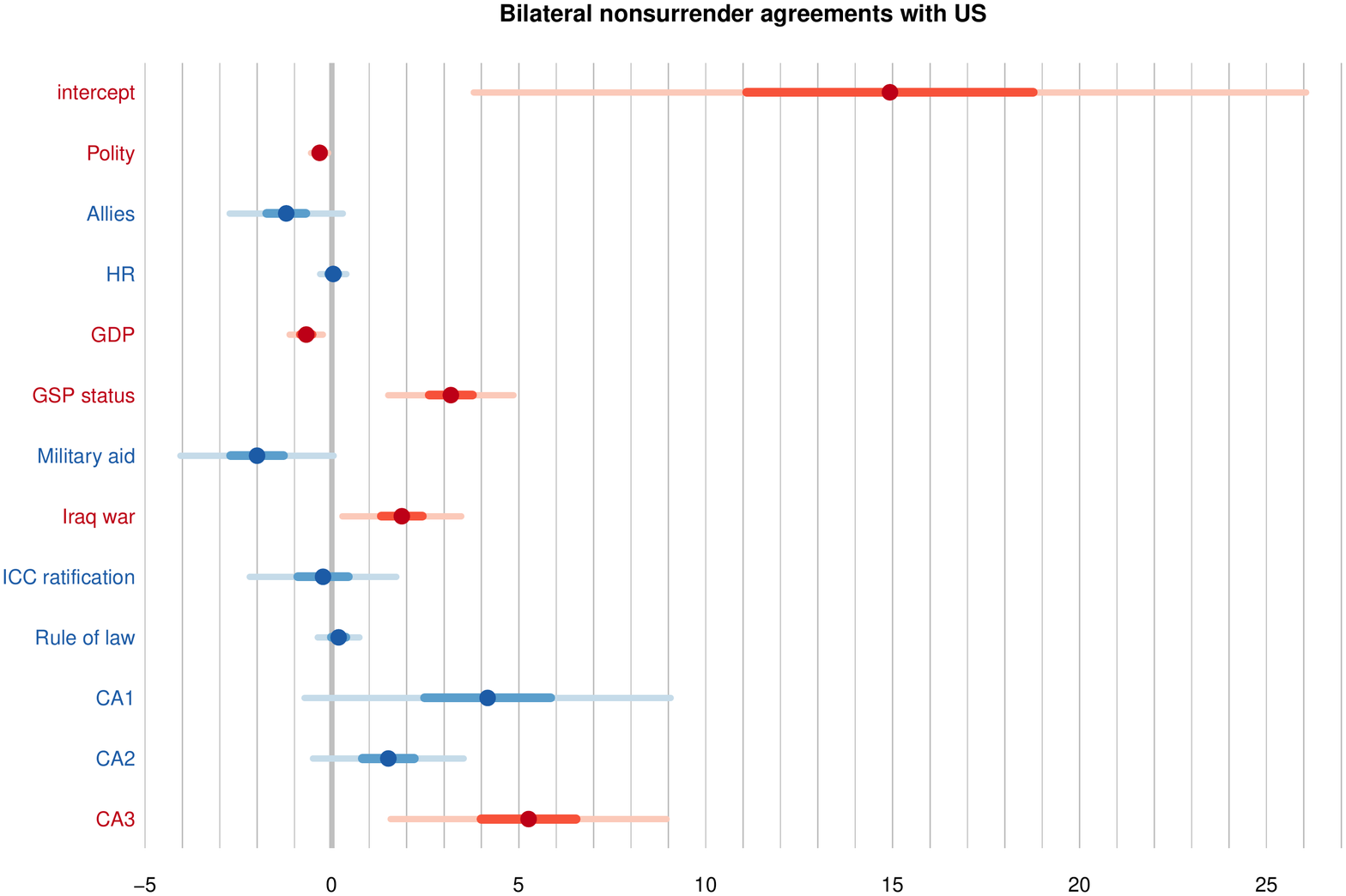}

\caption{\emph{Choosing Optimal Number of CA Dimensions and the Estimated Model Results}. \emph{Note:} The upper subplot displays the results of leave-one-out-cross-validation (LOOCV) analysis to choose the optimal number of dimensions for the estimated model, as discussed in the text. The bottom subplot displays the coefficients of the estimated model, as discussed in the text, with 50\% and 95\% CIs. Coefficients in red are statistically significant (at 95\% level) The specification has fewer observations than the original analysis as \cite{kelley2007keeps} includes non-UN members or states that did not participate in the GD that year, i.e., with absent text data.
\label{fig:cv}}

\end{figure}

To determine the optimal number of dimensional estimates to include in the estimation we rely on the leave-one-out cross-validation (LOOCV) method \citep[p. 178]{james2013introduction}. Given the sample size, we considered alternative specifications with up to ten CA dimensions, as presented in Figure \ref{fig:cv}.\footnote{\normalsize We implement the simplest specification search using additive models. Users can implement more extensive searches using a similar approach, e.g., including interaction terms.} For each alternative model we calculate the cross-validation error. As the LOOCV indicates that three CA dimensions is optimal, we include three dimensions to the original specification that predicts whether countries signed nonsurrender agreements \citep{kelley2007keeps}.

\begin{figure}
\centering

\includegraphics[width=.95\textwidth]{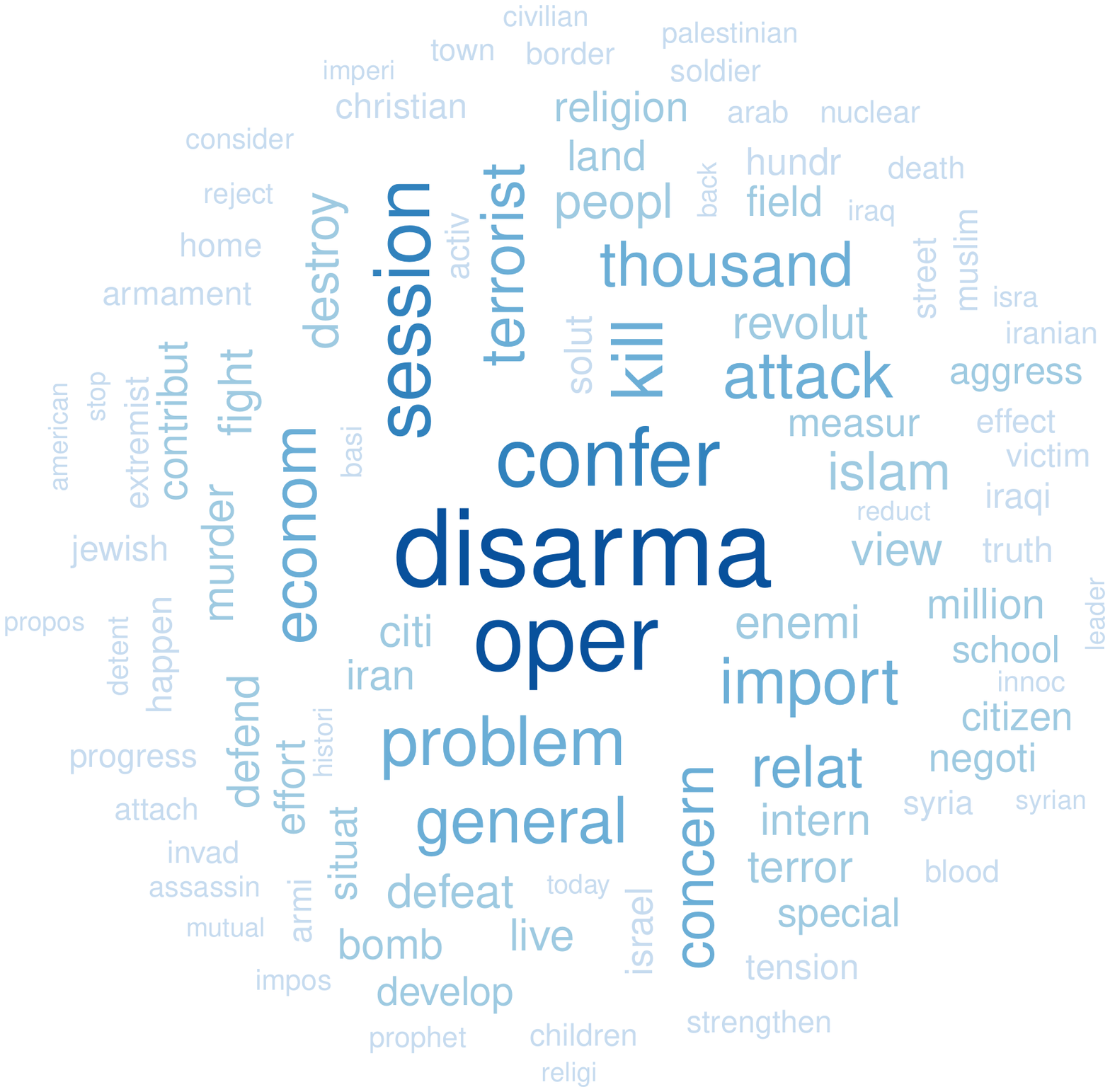}

\caption{\emph{Word Cloud of Top 100 Words on CA3}. \emph{Note:} 100 words are included that have the highest loading on CA3; the font size is proportional to a word's contribution to the variation on this dimension.
\label{fig:wordcloud}}

\end{figure}

The results presented in the second subplot in Figure \ref{fig:cv} indicate that the CA3 coefficient is statistically significant. What does this mean substantively? A detailed discussion is limited by the scope of the paper, but we can gain some insight from Figure \ref{fig:wordcloud}, which shows the most important words defining the variation on that dimension. The results suggest that states that expressed stronger concerns about security and terrorism were more likely to sign the nonsurrender agreement. We interpret this as indicating security concerns alongside normative goals influenced decisions on whether to sign the nonsurrender agreement with the US. It is, however, important to note that further analysis would be required to fully support this claim.

\vspace{-0.2in}
\subsection*{Conclusion}

This paper introduces a new dataset, the UNGDC, for understanding and measuring state preferences in world politics. We have demonstrated how scholars can extract relevant information from the UNGDC using text analytic methods. Specifically, we have shown how the UNGDC can be used to uncover single and multiple dimensions of government preferences, and have provided examples of how such estimates can be applied. 

Estimates derived from the UNGDC complement existing measures of government preferences based on UNGA voting. In fact, a possible application of the UNGDC would be to investigate the relationship between preferences expressed by governments in their GD statements and their voting behavior in the UNGA across difference issue areas. This would shed light on whether governments express their foreign policy preferences in different ways depending on the particular audience they face and the associated costs.  

A benefit of using texts to derive information about preferences is that they provide detailed information about countries' views on a particular policy area, and so can be compared to other text data. Hence, a future application of the UNGDC would be to compare the statements with international treaties and laws. Such comparisons can show whether some countries have greater influence on specific international agreements, and how countries perceive such agreements. For example, researchers may consider the extent to which states adopt language based on international law in their GD statements. Finally, in addition to examining the effects of government preferences, the UNGDC can also help us better understand how state preferences are formed, and which groups in a country influence preferences across different issues.

\end{spacing}

\clearpage  \singlespacing
\setstretch{1.1}
\theendnotes
\clearpage \singlespacing
\bibliographystyle{apsr}
\bibliography{un}

\end{document}